%% file: ms.tex
\documentclass{siamart190516}

\usepackage[utf8]{inputenc}
\usepackage{caption}
\usepackage{subcaption}
\usepackage{graphicx}
\usepackage{amsmath}
\usepackage{amssymb}
\usepackage{booktabs}
\usepackage{algorithm,algorithmicx,algpseudocode}
\usepackage{wrapfig}
\usepackage{bbm}
\usepackage{multirow}
\usepackage{enumitem}

\newcommand{\bE}{\ensuremath{\mathbb{E}}}

\newcommand{\SSIM}{\ensuremath{\text{SSIM}}}

\newcommand{\PieAPP}{\ensuremath{\text{PieAPP}}}
\input{math_commands.tex}

\newcommand{\shortparagraph}[1]{\noindent\textbf{#1}\quad}
\newcommand{\etal}{\textit{et al}.}
\usepackage{hyperref}
\begin{document}

\title{Human Imperceptible Attacks and Applications to Improve Fairness}

\author{Xinru Hua\thanks{Stanford University, Stanford, CA (\email{huaxinru@stanford.edu})} \and Huanzhong Xu\thanks{Stanford University, Stanford, CA (\email{xuhuanvc@stanford.edu})} \and Jose Blanchet\thanks{Stanford University, Stanford, CA (\email{jose.blanchet@stanford.edu})} \and Viet Nguyen\thanks{VinAI,
Vietnam (\email{vietanh.nguyen226@gmail.com})}}

\maketitle

\begin{abstract}
Modern neural networks are able to perform at least as well as humans in numerous tasks involving object classification and image generation. However, small perturbations which are imperceptible to humans may significantly degrade the performance of well-trained deep neural networks. We provide a Distributionally Robust Optimization (DRO) framework which integrates human-based image quality assessment methods to design optimal attacks that are imperceptible to humans but significantly damaging to deep neural networks. Through extensive experiments, we show that our attack algorithm generates better-quality (less perceptible to humans) attacks than other state-of-the-art human imperceptible attack methods. 
Moreover, we demonstrate that DRO training using our optimally designed human imperceptible attacks can improve group fairness in image classification. Towards the end, we provide an algorithmic implementation to speed up DRO training significantly, which could be of independent interest.
\end{abstract}

\section{Introduction}\label{sec:intro}
Deep learning models are making strides into our daily life with tremendous successes in diverse areas of applications, such as self-driving cars and face recognition. However, we still lack a fundamental understanding of how deep neural networks (DNNs) perceive and process information. One behavior of DNNs that we do not fully understand is how they are impacted by adversarial attacks. The potential implication of these attacks involves threats in, for instance, safety and robustness. Adversarial attacks provide a method to study and understand the relationship between machine perception and human perception. Over the years, neural network design has been inspired by the ways in which the human brain responds to visual stimuli~\cite{xu2021limits, voulodimos2018deep}. Although adversarial attacks are intended for DNNs, they may cause differences to human vision systems as well~\cite{zhou2019humans,elsayed2018adversarial}. In our work, we study adversarial attacks that are designed to primarily affect machine perception. We demonstrate that by training against human imperceptible attacks, we can improve the quality of the classification models in attributes that are interesting from a human perception standpoint, such as fairness.

A formal definition of adversarial attacks on image classification~\cite{Moosavi-Dezfooli_2016_CVPR} is the following. Given a classifier $f$, an image $\vx$, and a cost function $c$ on the image space, an optimal adversarial attack solves $\vDelta$ that can change the model's classification results via the smallest budget:
\begin{equation}
     \min_{\vDelta} c(\vx,\vx+\vDelta),\quad \text{with}\quad f(\vx+\vDelta)\ne f(\vx).
\label{eq:adv_attack}
\end{equation}
Traditional adversarial attack methods use $L_p$ distances as the cost function, see ~\cite{goodfellow2014explaining,madry2017towards,Moosavi-Dezfooli_2016_CVPR,tramer2018ensemble}. However, as reported in recent literature~\cite{sharif2018suitability, wang2004image}, $L_p$ distances do not accurately measure differences in human perception. One goal of this paper is the systematic study of adversarial attacks which are imperceptible to humans. 

\begin{figure*}[t]
\centering
\includegraphics[width=0.7\textwidth]{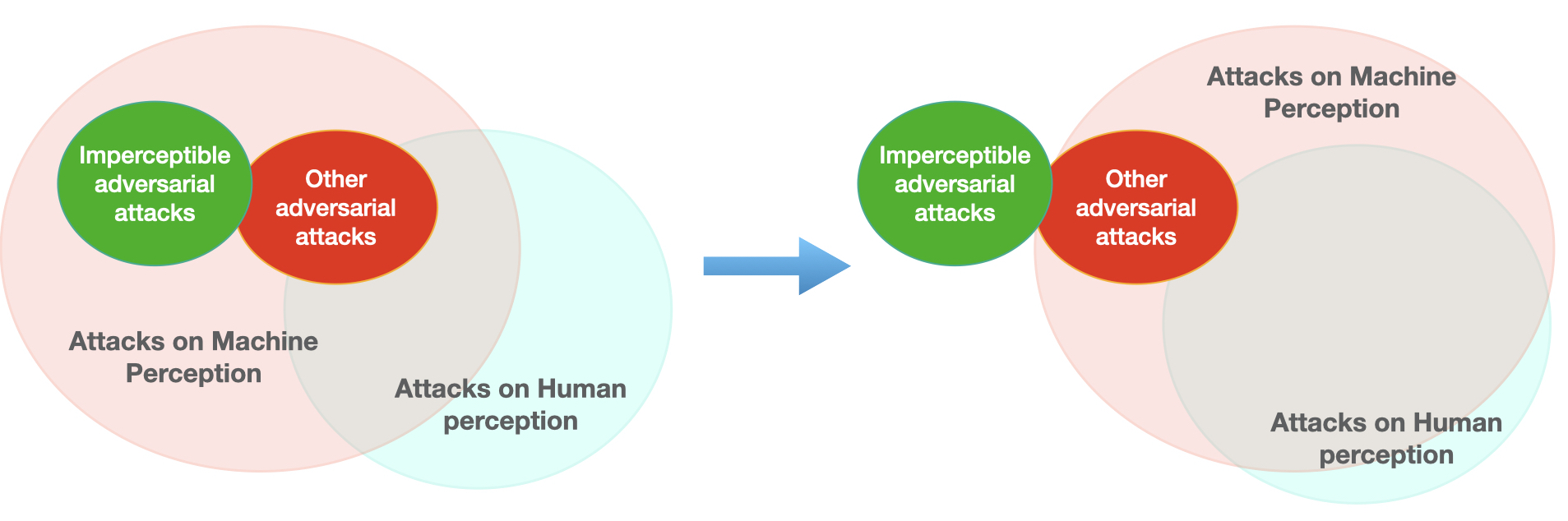}
\caption{\textbf{Left}: The two big circles represent the attacks or small perturbations that can cause changes visually in machine perception and human perception respectively. Commonly used adversarial attacks (represented as the red circle), for example, $L_p$ based adversarial attacks, are visually perceived by both machines and humans. We design our attacks (represented as the green circle) to be imperceptible to humans and only affect machine perception. Then we start adversarial training with our attack method and move the relationship as displayed in the left image to the right image. \textbf{Right}: Adversarial training with our attacks discourages the model perceiving the attacks in the green circle, so it pushes the machine perception circle in the direction of human perception. At the end of the adversarial training, the perturbations that machines perceive and humans perceive overlap more, so that the two perception systems align more closely.}
\label{fig:machine_human}
\end{figure*}

To design adversarial attacks that humans cannot perceive, the choice of cost function $c$ in~\cref{eq:adv_attack} is important. As mentioned before, $L_p$ cost functions may not constrain adversarial attacks to be imperceptible to humans. In our work, we constrain the attacked images to be close to the original image in two choices of cost function which measure human perceptual distances: structural similarity index measure (SSIM)~\cite{wang2004image} and PieAPP~\cite{prashnani2018pieapp}. These human perceptual distances are introduced by works in image quality assessment (IQA) methods. By qualitative and quantitative comparison, we show that our attacked images have less human perceptual differences to the original images than other state-of-the-art (SOTA) works. Moreover, we aim to align machine perception with human perception by using the Distributionally Robust Optimization (DRO)
framework incorporated with our attack method and we hypothesize that the aligning process reduces the models' biases. The relationship between our human-imperceptible attacks, other adversarial attacks, machine perception, and human perception is illustrated in~\cref{fig:machine_human}. The two images from left to right demonstrate how DRO training with our attack method aligns machine perception with human perception.

Recently, the DRO framework has been studied extensively in machine learning, because it can be used to compute the most reliable model under distributional uncertainty~\cite{blanchet2018optimal, rahimian2019distributionally}. DRO-trained models are able to achieve uniform performance across all groups of data, even on out-of-sample data~\cite{blanchet2021sample,volpi2018generalizing}. The DRO framework has been used in the context of data-driven distances~\cite{9004785}. In our work, we incorporate IQA methods as the cost function to the DRO framework and solve the adversarial attacks and DRO trained models. As shown in~\cref{fig:machine_human}, by using DRO training with our adversarial attacks, we reduce the model's dependence on human imperceptible features, so the models will focus more on features that humans can perceive and think are important. We show that DRO training with our attack method reduces more biases of the models compared to DRO training using the Projected Gradient Descent (PGD) attack method~\cite{madry2017towards}, and thus improves fairness.

\begin{figure}[t]
  \begin{center}
    \includegraphics[width=0.3\textwidth]{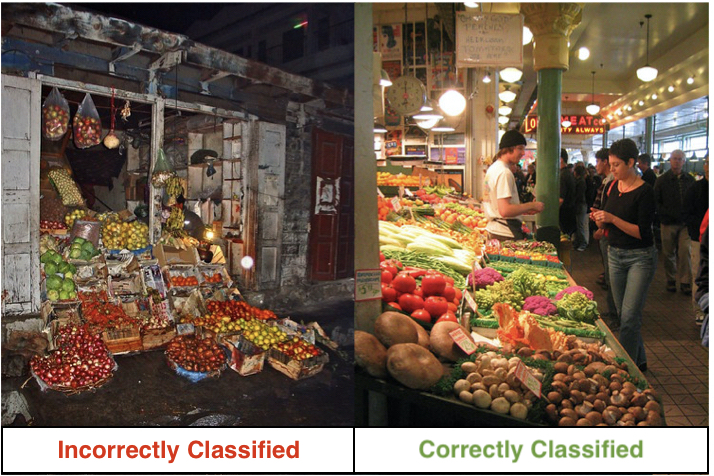}
  \end{center}
  \caption{Images from Pakistan (\textbf{Left}) and the US (\textbf{Right})}\label{fig:grocery}
\end{figure}

As DNNs achieve high performance in multiple tasks and we start to apply DNN models in daily applications, fairness has become more crucial, especially the question of whether the models perform equally well on the data from underrepresented groups. One of the crucial issues is that current datasets do not have a uniform distribution on images from all demographics and models may infer biases from unbalanced data. In both of the two popular open-source data sets: ImageNet and Open Images, approximately half of the images are collected from 2 countries: the United States and Great Britain~\cite{shankar2017no}. Moreover, DNNs are suspected to learn spurious features to help classification and the spurious features are learned from the majority groups~\cite{de2019does,khani2021removing,xiao2020noise}. Both works studying classification fairness~\cite{shankar2017no,de2019does} group images by the country where images are collected, so we follow the convention and collect our ImageNet geo-location dataset. 
\Cref{fig:grocery} shows two images of our dataset from the class grocery store, grocery, food market, market. The image from Pakistan is misclassified and the image from the US is not. We observe that images of grocery stores in higher-income countries are in relatively good lighting conditions and with a clean background, as opposed to images from less developed countries, which may be learned by the models as a spurious feature. By our proposed adversarial attack method and DRO training algorithms, we prevent the models from learning human imperceptible features, which are spurious features, and achieve more uniform performance on images from all groups.
Our work's contributions are summarized as follows:
\begin{enumerate}
    \item We connect human perceptual distances SSIM and PieAPP with the DRO framework to generate adversarial attacks that are imperceptible to humans and successfully attack classification models. We use methods introduced in the human vision learning area to show that our attacks are less perceptible to humans than other state-of-the-art imperceptible attacks. We add a confidence parameter to our algorithm, so our method with high confidence is the most successful method against two popular defense methods.
    \item We collect a dataset from ImageNet~\cite{ILSVRC15} with country information, and design two hypothesis tests to show that the original model has biases in classification and DRO training with our attacks reduces biases. The hypothesis tests can serve as a general methodology to test group fairness.
    \item We provide an algorithmic implementation of independent interest which can speed up DRO training. We add a few speed-up techniques to make generating attacks and DRO training more practical. 
\end{enumerate}

This paper unfolds as follows. In~\cref{sec:related}, we conduct a comprehensive literature review. In \cref{sec:method_main}, we introduce our adversarial attack method that computes optimal imperceptible attacks. We also show numerical comparison results and image comparisons. In \cref{sec:fairness}, we solve the DRO problem and compare fairness in DRO trained models with two adversarial attack methods. All the dataset, code and figures in full resolution are available in the supplementary material. 

\section{Related Work} \label{sec:related}
\shortparagraph{Adversarial attacks.} Since the seminal work of Ian Goodfellow \etal~\cite{goodfellow2014explaining}, there is a surge of papers studying adversarial attacks:~\cite{carlini2017towards,madry2017towards,Moosavi-Dezfooli_2016_CVPR,kurakin2016adversarial,dong2018boosting,chakraborty2018adversarial}. Some of the most notable papers use Wasserstein distance~\cite{wong2019wasserstein}, human perceptual distance~\cite{zhao2020towards,laidlaw2021perceptual}, attacks in feature space~\cite{xu2020towards}. Other non-conventional adversarial attacks are: sparse adversarial attacks~\cite{andriushchenko2020square, zhu2021sparse}, spatial perturbations~\cite{engstrom2019exploring,Zeng_2019_CVPR}, contour region attack~\cite{9527254}, and one-pixel attack~\cite{su2019one}. Other than white-box attack methods, there are many successful black-box attack methods~\cite{guo2019simple,ilyas2018black}. For a more detailed analysis on the adversarial attack literature, please refer to recent review papers~\cite{akhtar2018threat,chakraborty2018adversarial}.

\shortparagraph{Adversarial attack and human vision.} There is literature studying the influences of adversarial attacks on human vision~\cite{zhou2019humans, elsayed2018adversarial}. Zhao~\etal~\cite{zhao2017generating} generate adversarial attacks that are semantically meaningful, that can be perceived by humans. Madry~\etal~\cite{madry2017towards} also report that $L_2$ based attacks can be large enough to cause misclassification by humans.

\shortparagraph{Human perceptual distance.} We need distance functions to measure differences in human perception to truly restrict the size of adversarial attacks in human perception. Image quality assessment methods study how to measure human perceptual distance. Traditional IQA methods include SSIM~\cite{wang2004image}, MS-SSIM~\cite{1292216} FSIM~\cite{zhang2011fsim}, and PSNR~\cite{5596999}. DNN-based IQA methods include DISTS~\cite{ding2020iqa}, PieAPP~\cite{prashnani2018pieapp}, LPIPS~\cite{zhang2018perceptual}, PIM~\cite{bhardwaj2020unsupervised}, and SWD~\cite{gu2020image}.

\shortparagraph{Distributionally Robust Optimization (DRO).} As people care more about models' robustness in the extreme circumstances, DRO framework emerged to gain a lot of interests. There have been a number of theorectical work on DRO and Optimal Transport, see~\cite{blanchet2018optimal,blanchet2019quantifying, duchi2021learning,rahimian2019distributionally,kuhn2019wasserstein,staib2019distributionally,van2021data}. In particular, \cite{esfahani2018data, shafieezadeh2015distributionally, gao2016distributionally, gao2017wasserstein, blanchet2019robust} study the theory and applications of DRO problems using Wasserstein distance to parameterize the constraint set. \cite{volpi2018generalizing} generalizes models to unseen domains by training the models with DRO. \cite{sinha2018certifiable} first introduces combining DRO framework and adversarial attacks. \cite{yinpeng2020adv} introduces adversarial distributional training (ADT) to generalize models over the wrost adversarial distributions. Other adversarial training methods are introduced in \cite{tramer2018ensemble, zhang2019defense,madry2017towards}.

\shortparagraph{Fairness.}   Many recent papers discover unfairness in image classification and object detection models~\cite{de2019does, wilson2019predictive, buolamwini2018gender}. Specifically, these papers point out that neural network models discriminate against underrepresented groups. One possible explanatory factor of unfairness is that the open-source datasets are unbalanced~\cite{shankar2017no}. \cite{yang2019fairer,gong2012overcoming} starts to fix the datasets by collecting data that are representative among all demographics. In the natural language processing community, recent work discovers that word embedding models learn the biases from data~\cite{bolukbasi2016man,caliskan2017semantics}. A recent review on fairness in machine learning can be found in~\cite{mehrabi2021survey}.

\section{Method}\label{sec:method_main}
We consider the following DRO problem which finds the model that performs well in the worst-case distribution: 
\begin{equation}\label{eq:raw}
    \min_{\theta}\sup_{P:D(P,P_0)\le\delta}\E_P[\ell(\theta; X, Y)],
\end{equation}
where $\theta$ is the model parameter, $\ell$ is the pre-specified loss function. Distribution $P_0$ is the empirical distribution of $(X,Y)$ constructed from the training data, $D$ is a distance metric between probability distributions, and $\delta$ is the size of the distributional uncertainty. Similar to~\cite{sinha2018certifiable}, we choose the Wasserstein distance as our metric $D$. Specifically, let $c((\vx, y), (\vx', y'))$ denote the cost function to measure the distances between two training samples $(\vx, y)$ and $(\vx', y')$, and $\Gamma(P, P_0)$ denote the set of all joint distributions of $P$ and $P_0$, then our distance metric is given by
\begin{equation*}
    D(P, P_0) = \inf_{\gamma \in \Gamma(P, P_0)} \E_{\gamma} [c((\vx, y). (\vx', y'))],
\end{equation*}
 Here we use a separable ground cost $c((\vx, y), (\vx', y')) = c_0(\vx, \vx') + \infty \cdot \mathbbm{1}\{y \neq y'\}$, which penalize infinitely the perturbation of the labels. By~\cite[Theorem~1]{blanchet2019quantifying}, the DRO problem~\cref{eq:raw} is equivalent to
\begin{align}\label{eq:DRO}
    \min_{\theta}\inf_{\lambda\ge 0}\left(\lambda\delta+\E_{P_0}[\phi_{\lambda}(\theta;\vx_0,y_0)]\right),
\end{align}
where the robust surrogate loss $\phi_\lambda$ is defined by
\begin{equation}\label{eq:inner_sup}
    \phi_{\lambda}(\theta; \vx_0, y_0) = \sup_{\vx}~\ell(\theta;\vx, y_0)-\lambda c((\vx,y_0),(\vx_0,y_0)).
\end{equation}

Note that $\{\vx_0,y_0\}$ is one data sample, where $\vx_0\in \mathbb{R}^n$ is a vector representation of the image and $y_0$ is its label.

As discussed in \cref{sec:intro}, instead of using $L_p$ distances, we will define $c_0$ by two distances that better represent human perceptual distances, $1-\SSIM$ and PieAPP, which are discussed below in more detail:
\begin{enumerate}[leftmargin=5mm]
    \item $c_0(\vx,\vx') = 1-\SSIM(\vx,\vx')$. SSIM~\cite{wang2004image} is a reward function on two grayscale images that captures structural similarity in the two images. \cite{wang2009mean} shows a table of the same image distorted by different methods. The table of images demonstrates that, from the human perspective, two images with similar MSE errors have drastically different qualities, but two images with similar SSIM values are of a similar quality. SSIM's formula is stated in the supplementary material~\cite{supplementary}. Since $\SSIM\le 1$, we use $1-\SSIM$ as our cost function. $1-\SSIM$ satisfies three properties: symmetry, boundedness, and unique minimum~\cite{wang2009mean}, which makes it very close to a distance function.
    \item $c_0(\vx,\vx') = \PieAPP(\vx,\vx')$. PieAPP~\cite{prashnani2018pieapp} uses a DNN to measure two images' visual differences in human judgment. PieAPP can be applied on images of any size larger than $64\times64$. PieAPP's uniqueness is its novel pairwise preference probability, for instance, the probability that humans think one image is more of similar quality than another image when compared to a reference image. Pairwise comparison is more robust because humans may have clear preferences between all pairs of images but cannot assign scores to all images. Another advantage is that PieAPP does not depend on any existing architectures or pretrained models, as opposed to LPIPS and DISTS. 
\end{enumerate}
We solve problem~\eqref{eq:inner_sup} to obtain optimal adversarial attacks, which we discuss in the current section, and we solve the optimization problem~\eqref{eq:DRO} to obtain a distributionally robust model, which is in more detail in~\cref{sec:fairness}.
\subsection{SSIM-based attack}\label{sec:SSIM}
\begin{figure}[h]
\centering
\begin{subfigure}{0.25\textwidth}
\centering
\includegraphics[width=\textwidth]{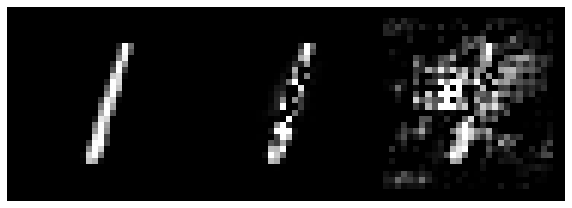}
\subcaption{The classified labels are: 1, 8, 4.}
\end{subfigure}
\begin{subfigure}{0.25\textwidth}
\centering
\includegraphics[width=\textwidth]{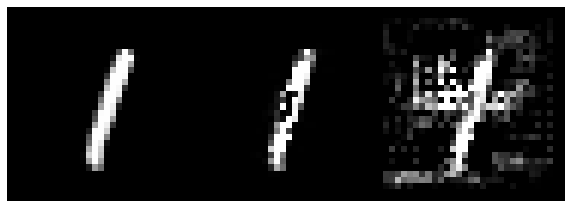}
\subcaption{The classified labels are: 1, 4, 4.}
\end{subfigure}
\caption{\textbf{Left}: Original image, \textbf{Middle}: Our one-step attack, \textbf{Right}: One-step PGD attack ($L_2$). We compare our one-step method and one-step PGD attack using $L_2$ cost function, both with $\eps=10$. Here are the first two 1's in MNIST test split that are successfully attacked by both methods. Our attack does not change the structure nor the true class of the numbers, but $L_2$ attacks make the digits unrecognizable and more similar to the mis-classified label. 
More examples are in the supplementary material~\cite{supplementary}.}\label{fig:SSIM}
\end{figure}

In this section, we focus on $c_0 = 1 - \SSIM$. Let $\mH_{\vx}(\vx')=\nabla_{\vx'}^2 c_0(\vx,\vx')$, which is $c_0$'s Hessian matrix with respect to the second argument. Given an input $(\vx_0, y_0)$, we use a one-step method to solve problem~\eqref{eq:inner_sup}:
\begin{align}\label{eq:one_step}
\begin{split}
 \vx_\textnormal{adv}&= \vx_0+\epsilon \frac{\vDelta^*}{\|\vDelta^*\|_2},\\
\text{with} \quad \vDelta^* &=  \mH_{\vx_0}(\vx_0)^{-1}\nabla_{\vx} \ell(\theta;\vx_0,y_0).
\end{split}
\end{align}
The derivation of \cref{eq:one_step} is given in the supplementary~\cite{supplementary}. Notice that our attack utilizes the Hessian of the cost function, which is different from the one-step PGD attack:
\begin{align*}
 \vx_\textnormal{pgd}&= \vx_0+\epsilon \frac{\nabla_{\vx} \ell(\theta;\vx_0,y_0)}{\|\nabla_{\vx} \ell(\theta;\vx_0,y_0)\|_2}.
\end{align*}
An illustrative comparison can be found in~\cref{fig:SSIM}.

\begin{figure*}[t]
\centering
\begin{subfigure}{\textwidth}
\centering
\includegraphics[width=0.9\textwidth]{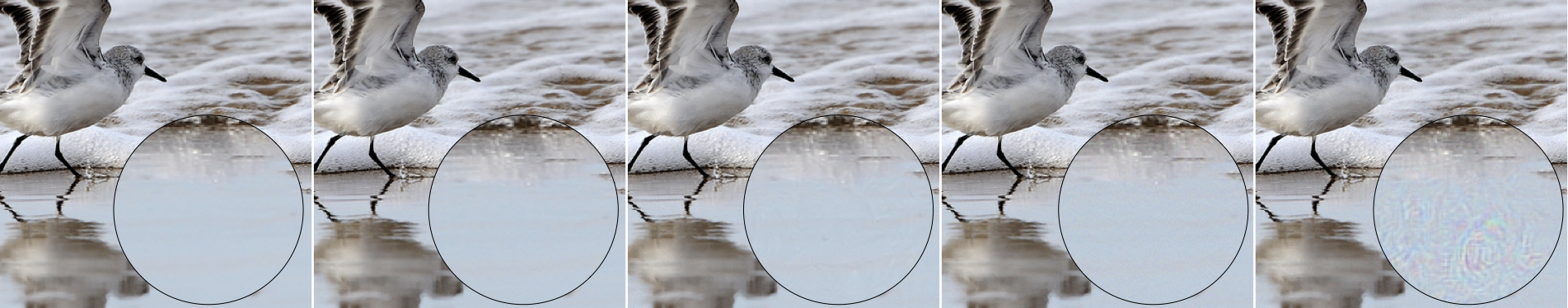}
\end{subfigure}
\centering
\begin{subfigure}{\textwidth}
\centering
\includegraphics[width=0.9\textwidth]{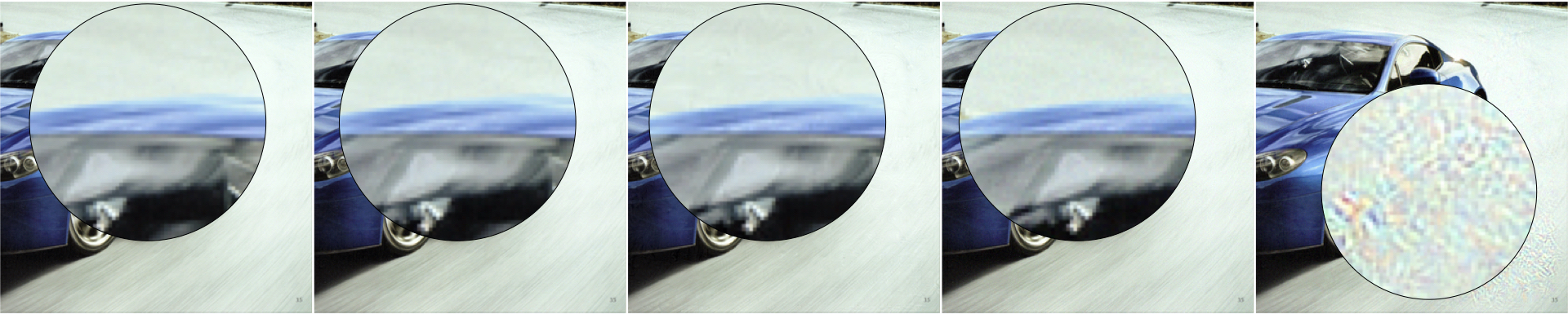}
\end{subfigure}
\begin{subfigure}{\textwidth}
\centering
\includegraphics[width=0.9\textwidth]{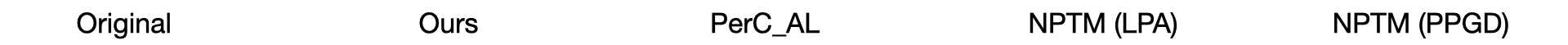}
\end{subfigure}
\caption{The comparison between the original image and adversarial attacks. Our method and PGD method generate images of similar visual quality, so we do not put PGD attack images here. We zoom in the regions where we can notice differences. We can see the PerC\_AL images have noticeable marble effects in both images. Both LPA images have noticeable sandy noises compared with other images and both PPGD images have an area of noises. The original images in full resolution and additional examples are in the supplementary material.}\label{fig:image_diff}
\end{figure*}

Due to the size and computation cost of the Hessian matrix, the above method is only practical with small images, for example on MNIST images of size $1\times28\times28$. We use this simple example to show that using $1-\SSIM$ as our cost function does discourage any structural changes that may change the true meaning of the images.
\subsection{PieAPP-based attack}
In this section, our choice of human perceptual distance $c_0$ is PieAPP. We attack images of size $3\times299\times299$ and use gradient descent to solve problem~\eqref{eq:inner_sup}, as described in~\cref{alg:attack_an_image}. We incorporate a confidence parameter in our early-stop mechanism to enhance the strength of our attacks, shown in line 5 in~\cref{alg:attack_an_image}, so they successfully attack defended images. We choose a ResNet-50 model pretrained on ImageNet~\cite{he2016deep} as the model $\theta$, cross-entropy loss as the loss function $\ell$, $N=100$, $\eps=0.1$, $\lambda=1$ and confidence $a=\{0,1,5\}$. We conduct ablation studies on the choices of parameters $\lambda,\eps$ in the supplement~\cite{supplementary}. 
\begin{algorithm}[htb]
    \caption{Attack an image}
    \label{alg:attack_an_image}
\algrenewcommand\algorithmicrequire{\textbf{Input:}}
\algrenewcommand\algorithmicensure{\textbf{Output:}}
\algnewcommand{\LineComment}[1]{\State \(\triangleright\) #1}
\algorithmicrequire{ image $\vx$, label $y$, classification model $\theta$, loss function $\ell$, confidence $a$, cost function $c_0$, number of iterations $N$,  step size $\epsilon$ }

\algorithmicensure{ adversarial image $\vx_\textnormal{adv}$}
    \begin{algorithmic}[1]
    \State \textbf{Initialize:}{ $\vx_\textnormal{adv} \leftarrow \vx$}
    \For{$k = 1,2,\ldots, N$}
    \State $\emph{logits} = \theta(\vx_\textnormal{adv})$
    
    \LineComment{$\emph{logits}_i$ means the logits before softmax for class $i$}
    \If {$\max_{i} \emph{logits}_{i\ne y}-\emph{logits}_y>a$} 
    \State \Return $\vx_\textnormal{adv}$ 
    \EndIf
    \State $\vDelta = \frac{\partial \ell(\theta; \vx_\textnormal{adv},y)}{\partial \vx_\textnormal{adv}}-\lambda \frac{\partial c_0(\vx,\vx_\textnormal{adv})}{\partial \vx_\textnormal{adv}}$ 
    \State $\vx_\textnormal{adv} \leftarrow \vx_\textnormal{adv}+\epsilon\vDelta$
    \State Validate $\vx_\textnormal{adv}$ \Comment{Validate $\vx_\textnormal{adv}$ as a RGB image}
    \EndFor
    \State \Return $\vx_\textnormal{adv}$
    \end{algorithmic}
\end{algorithm}

We compare our method with PGD ($L_2$), and with SOTA methods NPTM~\cite{laidlaw2021perceptual} and PerC~\cite{zhao2020towards} in terms of total time, success rate and human perceptibility\footnote{Hyunsik \etal~\cite{9527254} is another SOTA work, but its code is not public, so we do not compare with this work.}. We compare with PGD methods with 100 iterations and early-stop mechanism, and the formulation is \begin{align}
\nonumber
    \vx^0 &= \vx_0,\\
    \vx^{n} &= \vx^{n-1}+\epsilon \frac{\nabla_{\vx} \ell(\theta;\vx^{n-1},y_0)}{\|\nabla_{\vx} \ell(\theta;\vx^{n-1},y_0)\|_2}.
\end{align}

When comparing with the SOTA methods, we compare with NPTM (PPGD) and NPTM (LPA), the two main methods in the NPTM paper, and PerC\_AL, the faster and less perceptible method in PerC paper. Different from PerC and NPTM, our attack method directly solves the inner optimization problem~\eqref{eq:inner_sup}. PerC\_AL alternates between the two goals of attacking the image successfully and minimizing the perceptual distance, while our method combines the two goals in a single step. NPTM (PPGD) and NPTM (LPA) require an extra projection step, while our method does not.

Our attack method is evaluated on the development set of the\\ImageNet-Compatible dataset, which is the same as~\cite{zhao2020towards}. Since the dataset has 1000 images and we plan to compare against four other methods, involving humans to judge every pair of images is expensive. Thus, we apply two human perceptual distances and one salient object detection network to measure the differences in human perception induced by attacks. Note that the two human perceptual distances do not include PieAPP, because our method minimizes PieAPP distance and we want to objectively compare our method with other methods. We also show images to qualitatively judge the attacks' imperceptibility. 

\begin{figure*}[t]
\centering
\begin{subfigure}{0.1\textwidth}
\centering
\includegraphics[width=\textwidth]{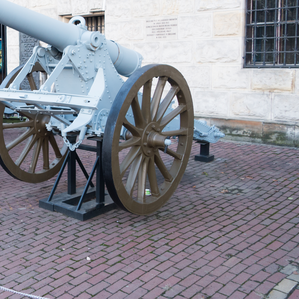}
\end{subfigure}
\centering
\begin{subfigure}{0.6\textwidth}
\centering
\includegraphics[width=\textwidth]{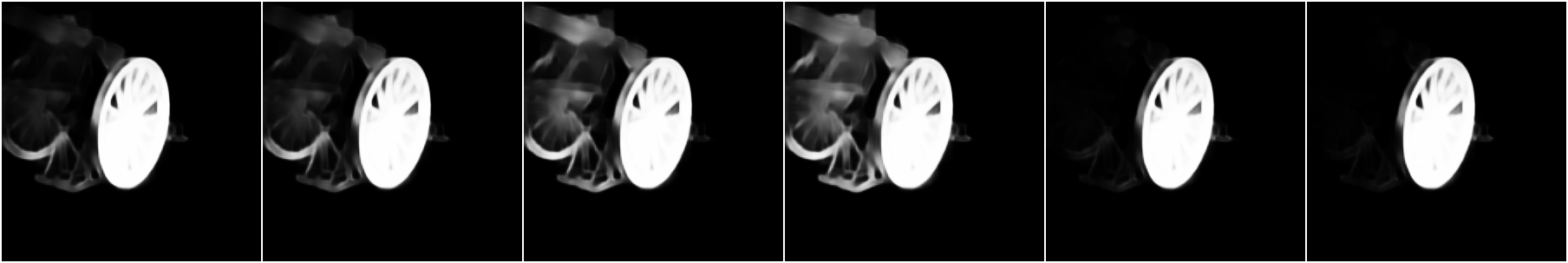}
\end{subfigure}\\
\begin{subfigure}{0.1\textwidth}
\centering
\includegraphics[width=\textwidth]{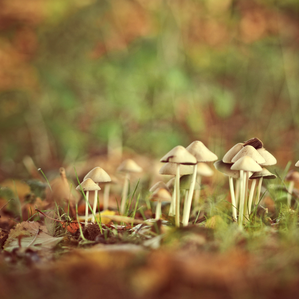}
\end{subfigure}
\begin{subfigure}{0.6\textwidth}
\centering
\includegraphics[width=\textwidth]{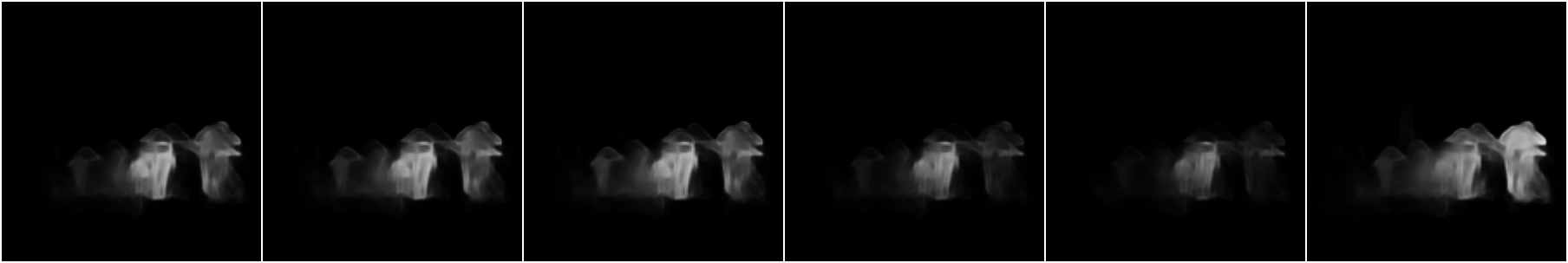}
\end{subfigure}
\begin{subfigure}{0.69\textwidth}
\centering
\includegraphics[width=\textwidth]{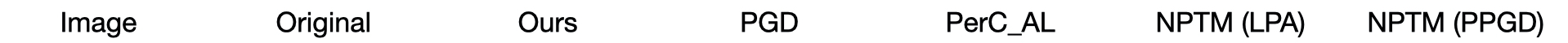}
\end{subfigure}
\caption{Comparison between the human saliency maps of the original image and attacked images. Compared with the original saliency map, our method generates the map with the least distinction. In the second image, NPTM (PPGD) shifts the attention from the mushrooms in the middle to the mushrooms in the right. More examples can be found in the supplementary material~\cite{supplementary}.}\label{fig:sal_diff}
\end{figure*}

First, we apply two Image Quality Assessment (IQAs) methods, LPIPS~\cite{zhang2018perceptual} and DISTS~\cite{ding2020iqa}, to quantify the perceptual distance between two images in human vision. The numerical results are given in \cref{table:image_diff}. \Cref{fig:image_diff} provides two examples to visually compare the quality of attacks. Finally, we apply EGNet~\cite{zhao2019EGNet} to images. EGNet is a model to predict a human saliency map, which means the object in an image that draws attention the most. We compute the human saliency maps of original images and the attacked images, and compute multiple distances between the original and attacked human saliency maps. \Cref{fig:sal_diff} illustrates two examples to qualitatively compare human saliency maps and~\cref{tab:human_saliency_diff} includes all the numerical comparison results.

\begin{table}[t]
    \centering \small
    {
  \centering \small
\resizebox{
  \ifdim\width>\textwidth
    0.8\textwidth
  \else
    0.82\width
  \fi
}{!}{%
\begin{tabular}{l |c|c c c c c@{}}
\toprule[1pt]
\multirow{2}{*}{\bf Approach}
&{\bf Success}
&\multicolumn{5}{c}{\bf Distance in adversarial images}\\
& \shortstack{\bf Rate (\%)}
& \shortstack{$L_1$}
& \shortstack{$L_2$}
& \shortstack{$L_\infty$}
& \shortstack{ LPIPS}
& \shortstack{ DISTS}
\\ 
\midrule[1pt]
PerC\_AL
& 100
& 633.12
& 2.22
& 0.085
& 33.96
& 33.82
\\
PGD ($L_2)$
& 100
& \bf{592.74}
& \bf{1.56}
& \bf{0.005}
& 7.82
& 8.77
\\
NPTM (PPGD)
& 95.75
& 2544.21
& 6.60
& 0.115
& 81.57
& 51.08
\\
NPTM (LPA)
& 99.78
& 2157.77
& 5.31
& 0.049
& 51.64
& 35.92
\\
\midrule
Ours ($a=0$)
& 100
& 783.86
& 1.91
& 0.006
& \bf{7.30}
& \bf{8.17}
\\
Ours ($a=1$) 
& 100
& 1965.10
& 4.45
& 0.014
& 22.24
& 21.47
\\
Ours ($a=5$)
& 100
& 3925.55
& 8.53
& 0.029
& 44.89
& 40.05
\\
\bottomrule[1pt]
\end{tabular}
}
\caption{\label{table:image_diff}The success rate is defined as the number of attacked images that labels change from being correct to incorrect divided by the number of correctly classified images. The distances measure the difference between attacked images and original images. The LPIPS and DISTS values are scaled by 1000. We embolden the smallest distance values in each column. Our method with $a=0$ has the \textbf{smallest} human perceptual distances (LPIPS and DISTS), despite larger $L_p$ distances than PGD.}
} 
\end{table}

\begin{figure}[t]
\centering
\begin{subfigure}{0.4\textwidth}
\centering
\includegraphics[width=\textwidth]{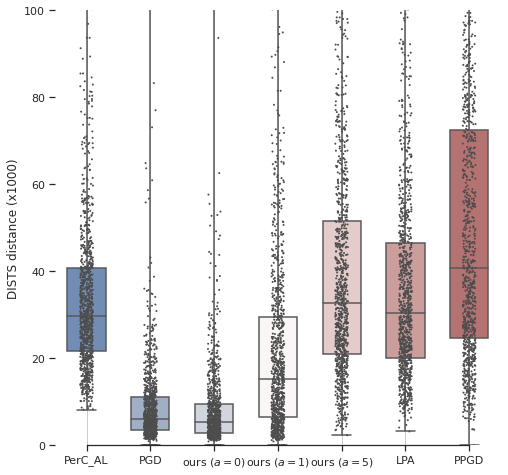}
\end{subfigure}
\begin{subfigure}{0.4\textwidth}
\centering
\includegraphics[width=\textwidth]{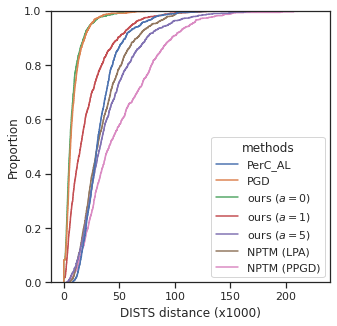}
\end{subfigure}
\caption{DISTS distances between the attacks and original images. \textbf{Top}: Boxplots of empirical DISTS distance distribution of all methods. The boxes display each distribution's 3rd quantile, median, and 1st quantile. Our method with $a=0$'s distribution has the \textbf{smallest} quantiles. \textbf{Bottom}: The cumulative distribution function of empirical DISTS distance distribution of all attack methods. For any DIST distance $d$, our method with $a=0$ has the \textbf{largest} $\probP(\vx_\textnormal{adv}\le d)$. Our method with $a=1$ has smaller perceptual distances than PerC\_AL, NPTM (LPA), and NPTM (PPGD). The plots for LPIPS distance are in the supplementary material.}
\label{fig:DIST_distance}
\end{figure}

\begin{figure}[H]
\centering
\begin{subfigure}{0.35\textwidth}
\centering
\includegraphics[width=\textwidth]{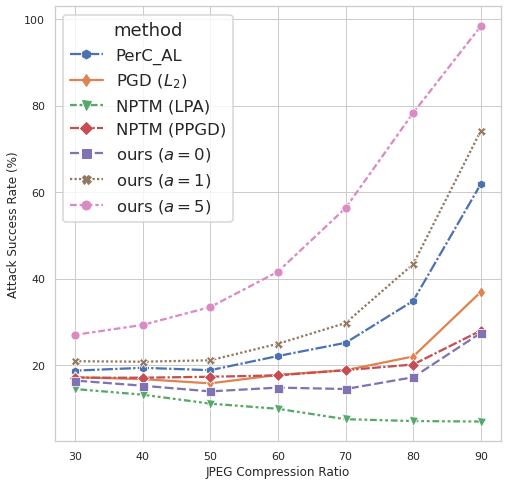}
\end{subfigure}
\begin{subfigure}{0.35\textwidth}
\centering
\includegraphics[width=\textwidth]{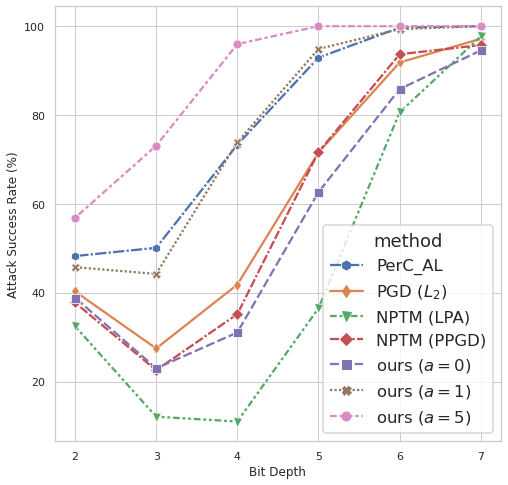}
\end{subfigure}
\caption{\textbf{Left}: JPEG compression ratio. \textbf{Right}: Bit Depth. After the attacked images being processed with JPEG compression defense and bit depth compression defense, our method with confidence 5 has the highest attack success rate in both defense methods and the images are still less perceptible than NPTM (PPGD)'s images (\cref{fig:DIST_distance}). Our method with confidence 1 generates less perceptible attack images than PerC\_AL and has a higher attack success rate in JPEG compression defense.}
\label{fig:comparisons}
\end{figure}

\begin{table}[t]
    {
  \centering \small
\resizebox{
  \ifdim\width>\textwidth
    0.9\textwidth
  \else
    0.9\width
  \fi
}{!}{%
\begin{tabular}{l|c| c c c c @{}}
\toprule[1pt]
\multirow{2}{*}{\bf Approach}
&{\bf Success}
&\multicolumn{4}{c}{\bf Distance in human saliency maps}\\
& \shortstack{\bf Rate (\%)}
& \shortstack{$L_1$}
& \shortstack{$L_2$}
& \shortstack{$L_\infty$}
& \shortstack{SSIM}
\\ 
\midrule[1pt]

PerC\_AL
& 100
& 1129.92
& 10.80
& 0.319
& 0.093
\\
PGD ($L_2)$
& 100
& 385.70
& 3.88
& 0.132
& 0.020
\\
NPTM (PPGD)
& 95.75
& 946.39
& 9.07
& 0.276
& 0.086
\\
NPTM (LPA)
& 99.78
& 522.03
& 5.34
& 0.188
& 0.036
\\
\midrule[1pt]
Ours ($a=0$) 
& 100
& \bf{325.73}
& \bf{3.34}
& \bf{0.118}
& \bf{0.016}
\\
Ours ($a=1$) 
& 100
& 654.57
& 6.42
& 0.205
& 0.045
\\
Ours ($a=5$)
& 100
& 1061.57
& 10.21
& 0.310
& 0.087
\\
\bottomrule[1pt]
\end{tabular}
}
 \caption{\label{tab:human_saliency_diff}The distances measure the differences between attacked images' saliency maps and original images' saliency maps. We embolden the smallest distance value. Our method with $a=0$ generates attack images with the \textbf{smallest} distances in the human saliency map, which means they induce the smallest changes in the human attentional mechanism.}
}
\end{table}

At last, we want to test our adversarial attack method's effectiveness against defense methods. Without knowing which adversarial attack is applied to the image, there are generic defense methods against attacks. We test our method on two such defense methods: \textbf{JPEG compression}~\cite{das2018shield,dong2019evading,dziugaite2016study,guo2017countering} and \textbf{bit depth reduction}~\cite{guo2017countering,he2017adversarial,xu2017feature}. The comparison with other methods is shown in~\cref{fig:comparisons}.
\section{Fairness}\label{sec:fairness}
In this section, we introduce methods to test whether a classifier suffers from biases and to measure their magnitude. Then we present two algorithms that can solve the DRO problem~\cref{eq:DRO} efficiently. We show that the training algorithms with adversarial attacks can mitigate the biases in models and training with our adversarial attack reduces biases more than with PGD attack. We are motivated by the fact that we can successfully attack a classifier by our human imperceptible attacks, so the classifier utilizes perturbations that humans cannot see and will not use in classification. By adversarial training with these human imperceptible perturbations, we prevent the model from using these attacks as classification features. In such a way, the model is less likely to learn human imperceptible features, which tend to be spurious features, and thus becomes fairer.

The nature of training DNNs method, which is minimizing the summation of loss function on all the images, means that models will favor features from the majority group, and this will degrade their performance on underrepresented groups. Recent works reveal that current open-source large datasets are severely unbalanced in the geographical location of images, and public object recognition systems do not perform uniformly on images from different locations~\cite{shankar2017no,de2019does}. Our collected ImageNet geo-location dataset is also amerocentric and eurocentric as discovered in~\cite{shankar2017no}, see a pie chart figure in the supplementary material. The detailed procedure of collecting the dataset is also in the supplementary material~\cite{supplementary}. 

Thus, we want to test whether a classifier has biases towards images from certain locations. Let $\mu$ be a probability model defined on the space $\mathcal{X}\times\mathcal{Y}$ of image and label pairs, and dataset $\mathcal{D}$ is a sample of $N$ independent and identically distributed (image, label) pairs, each following distribution $\mu$. Let $(\vx,y)$ be one test sample independent of the training set $\mathcal{D}$ following distribution $\mu$ and $Q$ be the joint distribution of the $N+1$ samples from $\mu$. We write $\bE$ as the expectation under $Q$. We use $\mathbbm{1}(\vx,y; \theta)$ to represent a Bernoulli event that model $\theta$ classifies $(\vx,y)$ correctly or not.

Given a fixed architecture with parameters $\theta$, we want to prove that our algorithm can improve classification fairness of $\theta$. For each image $\vx$ in $\mathcal{D}$, we can access its geographical location and the income level of the location, which is defined as $\rg(\vx)$. For a fair classifier and a random out of sample $\vx$, the classification accuracy should be \emph{independent} of the income level of the image:
\begin{equation}
     \E[\mathbbm{1}(\vx,y; \theta(\mathcal{D}))] = \E[\mathbbm{1}(\vx,y; \theta(\mathcal{D})) |\rg(\vx)].
     \label{eq:fairness}
\end{equation}
\Cref{eq:fairness} means the classification accuracy is uncorrelated with the income level of the image, which is a necessary condition for the independence to hold. Given this fairness definition, we group images by the country they are from, because each country is associated with one income level. Then we study if there exists a correlation between income level and classification accuracy of each country, which will be discussed in more detail.

We design \cref{alg:generate_adv} to generate an augmented dataset $\mathcal{D}_\textnormal{rob}$ and \cref{alg:train_adv} to approximate the solution to the DRO problem~\eqref{eq:DRO}. To conduct hypothesis testing, we run \cref{alg:train_adv} 50 times to sample 50 $\theta$'s based on $\mathcal{D}_\textnormal{rob}$ and the randomness of $\theta$ comes from the training process. More implementation details are in the supplementary material.

\begin{algorithm}[t]
    \caption{Generate an adversarial dataset $\mathcal{D}_{rob}$}
    \label{alg:generate_adv}
\algrenewcommand\algorithmicrequire{\textbf{Input:}}
\algrenewcommand\algorithmicensure{\textbf{Output:}}
\algorithmicrequire{ initial model $\theta_0$, learning rate $\alpha$, dataset $\mathcal{D}=\{x_i, y_i\}_{i=1,...N}$, number of steps $T_1$}

\algorithmicensure{ robust dataset $\mathcal{D}_\textnormal{rob}=\{x_i, y_i, P_i\}_{i=1,...M}$}
    \begin{algorithmic}[1]
    \State \textbf{Initialize:} $\theta = \theta_0, \mathcal{D}=\{x_i, y_i,P_i\}_{i=1,...N}$ with $P_i=1$
    \For{$k = 1,2,\ldots, T_1$}
    \State Sample $\{x_i, y_i, P_i\}_{i=1,...N}$ proportionally to the weights $P_i$ with replacement from dataset $\mathcal{D}$
    \For{$i = 1,2,\ldots, N$}
    \State $\theta \leftarrow \theta - \alpha P_i\nabla_{\theta}\ell(\theta;x_i,y_i)$
    \State Input $\theta, x_i, y_i$ to~\cref{alg:attack_an_image} to generate attack $\{x_i', y_i\}$
    \State Append $\{x_i', y_i, P_i\}$ to dataset $\mathcal{D}$ with weight $P_i=(k-1)N+i$
    \EndFor
    \EndFor
    \State \Return dataset $\mathcal{D}_{rob}$
\end{algorithmic}
\end{algorithm}
\begin{algorithm}[t]
    \caption{DRO training with a given adversarial dataset}
    \label{alg:train_adv}
\algrenewcommand\algorithmicrequire{\textbf{Input:}}
\algrenewcommand\algorithmicensure{\textbf{Output:}}
\algorithmicrequire{ initial model $\theta_0$, learning rate $\alpha$, robust dataset $\mathcal{D}_\textnormal{rob}=\{x_i, y_i, P_i\}_{i=1,...M}$, number of steps $T_2$}

\algorithmicensure{ DRO trained model: $\theta$}
    \begin{algorithmic}[1]
    \State \textbf{Initialize:} $\theta = \theta_0$
    \For{$k = 1,2,\ldots, T_2$}
     \For{$i=1,2,\ldots, M$}
      \State Sample $\{x_i, y_i\}$
     proportionally to the weights $P_i$ with replacement from dataset $\mathcal{D}_\textnormal{rob}$
    \State Set $\theta \leftarrow \theta - \alpha P_i\nabla_{\theta}\ell(\theta;x_i,y_i)$
    \EndFor
    \EndFor
    \State \Return model $\theta$
\end{algorithmic}
\end{algorithm}

 The intuition behind~\cref{alg:generate_adv} is that the outer loop chooses batches of size $N$ and the batches are sampled biased towards recent iterations. In turn, adversarial examples are added in the inner loop corresponding to the current optimization model parameters, which are updated according to standard stochastic gradient descent. The overall result is similar to a two-time-scale stochastic approximation algorithm, ~\cite{borkar1997stochastic}, which will be analyzed in future work. Our two algorithms' difference from the algorithm in~\cite{volpi2018generalizing} is that we can only run~\cref{alg:generate_adv} once and sample many models.

Given all the images and a model, we test the model on all images and compute accuracies by groups. We use a significance level of 0.05 in the following tests. We denote the groups as $\{\rvg_i, \rvp_i\}$, where $\rvg_i$ is the per capita GDP of $i$th country in log scale and $\rvp_i$ is the accuracy of classifying the images in $i$th country. We assume that the error in accuracy of each country is negatively related to the number of images in the country, so we write a covariance matrix as $\mSigma_{ii}=1/\sqrt{n_i}$, where $n_i$ is the number of images in the $i$th country. Then we run generalized least squares (GLS) with $\mSigma$ on data $\{\rvg_i, \rvp_i\}$ and obtain a linear estimator $\rvp=\beta \rvg+\eps$. 

\begin{figure}[t]
\centering
\includegraphics[width=0.45\textwidth]{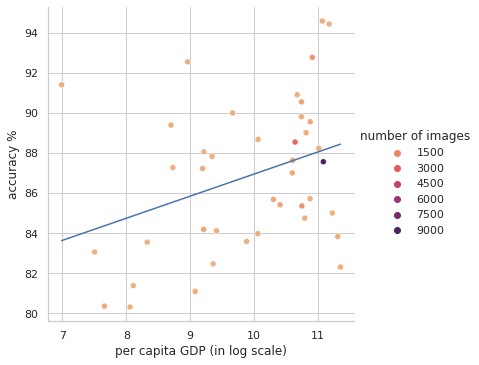}
\caption{We show a linear regression model with independent variable $\rvg$ and dependent variable $\rvp$, which are tested with the pretrained model $\theta_0$. Each dot represents one country and the color denotes the number of images in this country.}
\label{fig:pretrained_reg}
\end{figure}

Given the pretrained model $\theta_0$ and computed linear regression model $\rvp=\beta_0 \rvg+\eps$ (see~\cref{fig:pretrained_reg}), we hypothesis test whether there exists a non-zero linear relationship between $\rvp$ and $\rvg$ using the hypotheses:
\begin{equation*}
 H_0: \beta_0 = 0 \quad \text{versus} \quad H_1: \beta_0 \ne 0.
\end{equation*}
An F-test~\cite{hahs2020introduction} returns the computed score $F_0= 5.392$ with degrees of freedom $\nu_1=1, \nu_2=39$ and probability value $\probP(f>F_0)=0.02554$. Thus, we can reject the null hypothesis and conclude that there may exist a significant linear relationship between $\rvg$ and $\rvp$. For this pretrained model, group-fairness is not guaranteed.

\begin{figure}[t]
\centering
\includegraphics[width=0.3\textwidth]{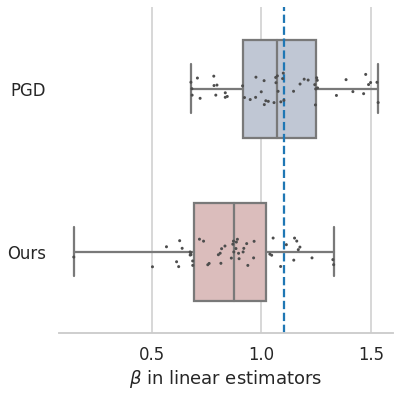}
\caption{The two boxes each shows the distribution of $\beta$ obtained from PGD method and our method respectively. The blue dashed line represents $\beta_0$ of the pretrained model. Both methods reduce the correlation between $\rvg$ and $\rvp$. Training with our attack method reduces biases more than PGD method.}
\label{fig:fairness}
\end{figure}

After we train 50 models and compute 50 linear estimators with our method and PGD method respectively, we use a standard t-test to compare the two means of $\beta$s~\cite{10.2307/41133999} to evaluate the magnitude of biases. We denote $\beta_m$ as our model's mean and $\beta_m'$ as PGD model's mean and conduct the hypothesis testing:
\[
    H_0: \beta_m \ge \beta_m' \quad \text{versus} \quad H_1: \beta_m < \beta_m'
\]
We compute the t-value is 3.852 with a probability $0.00017$, so we can reject the null hypothesis and conclude that $\beta_m<\beta_m'$. \Cref{fig:fairness} illustrates the two distribution of $\beta$'s. The pretrained model has an accuracy of 87.5\% on the test dataset. The models trained with our adversarial attack have a mean accuracy of 84.3\% and models trained with PGD method have a mean accuracy of 79.7\%. Thus, the models using our adversarial training method have a higher mean accuracy than the PGD method.
\section{Conclusion}
In this work, we present a method to generate human imperceptible adversarial attacks and design two DRO training algorithms. We show that our adversarial attack method can generate successful and least human perceptible attacks compared with other SOTA methods. For the ImageNet dataset and a model trained on it, we test the existence of inherent unfairness, such as geo-location biases. After testing two collections of models that are respectively trained by the DRO algorithm with our attack method and with the PGD attack method, our method improves fairness more significantly than the PGD method. Our hypothesis tests provide a general framework to test group fairness on the space of models conditioned on datasets. The limitation of our method is that we do not have enough computational resources or data to sample datasets, so we can only condition on one dataset and randomize the models. By generating a variation of adversarial attacks, our method mitigates the biases in the given dataset. We hope future work will incorporate the randomness in datasets and conduct the complete test in fairness. We also hope our work can help understand the differences between machine perception and human perception, and bridge the two areas of adversarial attacks and fairness in machine learning.

One negative use of our work is to maliciously apply imperceptible adversarial attacks to classification models. Solutions could be designing algorithms to detect imperceptible attacks and adversarial training algorithms to robustify models, which are future extensions to our work. The fairness application of our adversarial attacks brings good societal impacts, since we train the models to perform more uniformly on all data groups and thus become fairer.

\newpage
{\small
\bibliographystyle{ieee_fullname}
\bibliography{egbib}
}
\end{document}

%% file: math_commands.tex
\usepackage{amsmath,amsfonts,bm}

\def\1{\bm{1}}

\def\eps{{\epsilon}}

\def\rg{{\textnormal{g}}}

\def\rvg{{\mathbf{g}}}

\def\rvp{{\mathbf{p}}}

\def\vDelta{{\bm{\Delta}}}

\def\vx{{\bm{x}}}

\def\mH{{\bm{H}}}

\def\mSigma{{\bm{\Sigma}}}

\DeclareMathAlphabet{\mathsfit}{\encodingdefault}{\sfdefault}{m}{sl}
\SetMathAlphabet{\mathsfit}{bold}{\encodingdefault}{\sfdefault}{bx}{n}

\newcommand{\probP}{\text{I\kern-0.15em P}}
\newcommand{\E}{\mathbb{E}}

